\pdfoutput=1

\documentclass[11pt]{article}

\usepackage[]{acl}

\usepackage{times}
\usepackage{latexsym}
\usepackage{graphicx}
\usepackage{booktabs}
\usepackage{makecell}
\usepackage{algorithmic}
\usepackage{amsmath}
\usepackage[ruled,linesnumbered]{algorithm2e}
\usepackage[english]{babel}
\usepackage{graphicx}
\usepackage{hyperref}
\usepackage{caption}
\usepackage{multirow}
\usepackage{multicol}
\usepackage{paralist}
\usepackage{booktabs}

\usepackage[T1]{fontenc}

\usepackage[utf8]{inputenc}

\usepackage{microtype}

\usepackage{inconsolata}

\title{\textbf{\textit{Call Me When Necessary:}} LLMs can Efficiently and Faithfully Reason over Structured Environments}

\author{Sitao Cheng\textsuperscript{1}\thanks{This work is done during the internship at Microsoft.}, Ziyuan Zhuang\textsuperscript{1}\samethanks, Yong Xu\textsuperscript{2}, Fangkai Yang\textsuperscript{2}, Chaoyun Zhang\textsuperscript{2}, Xiaoting Qin\textsuperscript{2}, \\ \textbf{Xiang Huang\textsuperscript{1}, Ling Chen\textsuperscript{2}, Qingwei Lin\textsuperscript{2}, Dongmei Zhang\textsuperscript{2}, Saravan Rajmohan\textsuperscript{2}, Qi Zhang\textsuperscript{2}}\\ 
\textsuperscript{1}State Key Laboratory for Novel Software Technology, Nanjing University, China \\
\textsuperscript{2}Microsoft \\
\{stcheng, ziyuan.zhuang\}@smail.nju.edu.cn, \{yox, fangkaiyang\}@microsoft.com 
}

\begin{document}
\newcommand{\readi}{\textsf{Readi}}
\newcommand{\sota}{state-of-the-art}
\newcommand*\samethanks[1][\value{footnote}]{\footnotemark[#1]}

\maketitle
\begin{abstract}
Large Language Models~(LLMs) have shown potential in reasoning over structured environments, \textit{e.g.,} knowledge graphs and tables. Such tasks typically require multi-hop reasoning, \textit{i.e.,} match natural language utterance with instances in the environment.
Previous works adopt LLMs to incrementally build a reasoning path, where LLMs either invoke tools or pick up items by step-by-step interacting with the environment.
We propose \textbf{Re}asoning-P\textbf{a}th-E\textbf{di}ting (\readi), a novel framework where LLMs can efficiently and faithfully reason over structured environments.
In \readi, LLMs initially generate a reasoning path given a query,
and edit the path only when necessary.
We instantiate the path on structured environments and provide feedback to edit the path if anything goes wrong.
Experimental results on three KGQA and two TableQA datasets show the effectiveness of \readi, significantly surpassing previous LLM-based methods (by 9.1\% Hit@1 on WebQSP, 12.4\% on MQA-3H and 9.5\% on WTQ), comparable with \sota~fine-tuned methods (67\% on CWQ and 74.7\% on WebQSP) and substantially boosting the vanilla LLMs (by 14.9\% on CWQ). Our code will be available on \url{https://aka.ms/readi}.

\end{abstract}

\section{Introduction}
\label{intro}

Large Language Models (LLMs) 
have demonstrated remarkable performance in NLP fields~\cite{ouyang2022training,openai2023gpt4,liang2023holistic}. 
To further unleash their reasoning ability in complex scenarios, delicate strategies are proposed to equip LLMs with human-like thought process (\textit{e.g.,} Chain-of-thought~\citealp{wei2023chainofthought})
or leverage them as autonomous agents capable of planning, reflection and execution~\cite{yao2023react,liu2023agentbench}.
One compelling scenario where LLMs showcase their potential is reasoning over \textbf{structured environments (SEs)}~\cite{jiang2023structgpt,sun2023thinkongraph}. With dedicate schemas, SEs (\textit{e.g.,} knowledge graphs, tables) abstract real-world semantics for representing, storing, and querying data with relational structures.
For instance, Freebase~\cite{bollacker2008freebase} captures 45M entities and 3B facts over 100 domains, organized in triple patterns.
The crux of successful reasoning lies in bridging the gap between natural language and the mechanism of how the SEs are represented and operated~\cite{gu2023dont, li2023fewshot}.

\begin{figure}
    \setlength{\abovecaptionskip}{0.2 cm}
    \setlength{\belowcaptionskip}{-0.5cm}
    \centering
    \includegraphics[scale=0.5]{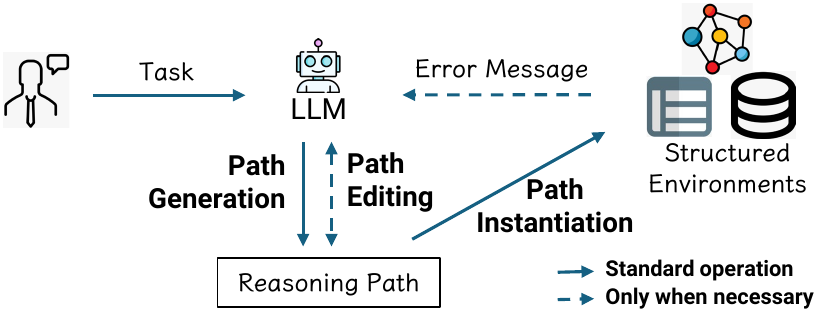}
    \caption{An illustration of our proposed framework, \readi,
    where LLMs initially generate a reasoning path, and when necessary, edit this path. We instantiate the path on structured environments and invoke editing if the instantiation gets stuck.
    }
    \label{fig:framework}
\end{figure}

While LLMs exhibit promising capabilities, their performance often falls short when faced with multi-hop reasoning involving large-scale SEs. 
To faithfully reason, prior works adopt an iterative way that start from certain elements (\textit{e.g.,} entity, relation in KG, column in Table), instantiate on SEs and then gradually expand the reasoning path while consuming just relevant portions instead of the entire environment
\cite{sun2023thinkongraph, jiang2023structgpt,huang2024queryagent}. This step-by-step interaction could mitigate hallucination of LLMs to some extent. However, the reasoning efficiency is sacrificed and thus hinder the practical feasibility.
For a simple constraint, say \texttt{``the daughter of Obama''}, such methods probably require two LLM-calls to first query the relations around \texttt{``Obama''} and then select \texttt{``father\_of''} from returned candidates. Moreover, at each step, LLMs make one choice based on history, making it prone to error propagation.
Alternatively, fine-tuned methods inject environments into model parameters by tuning with human-labeled supervision.
During inference, they recall schema patterns to build reasoning paths \cite{Zhang_2022_sr,saxena-etal-2020-improving,luo2023reasoning,ding2024enhancing,xiong2024large} without interaction with SEs. This end-to-end paradigm is efficient. However, it is never ensured that the model output can be grounded on SEs. Study shows that 50\% paths of RoG \cite{luo2023reasoning} failed to yield faithful results. In addition, they heavily rely on annotations, which are difficult to obtain for large-scale SEs.
Further research is thus needed to achieve efficient (\textit{i.e.,} less LLM-calls) and faithful (\textit{i.e.,} grounded on SEs) reasoning over massive SEs.

Therefore, we seek to propose an interaction paradigm that leverage LLMs to support complex reasoning on large-scale SEs faithfully and efficiently. Conducting empirical study on KGQA task, we find that 46\%-60\% of reasoning paths initially generated by LLMs can be well instantiated, which inspires us to fully exploit LLM’s intrinsic planning ability in complex reasoning.
While the idea of plan-and-refine is straightforward and applied in various real world tasks~\cite{taskweaver, embodimentcollaboration2023open,ahn2022i},
it’s worth noting that there exists few research on application in SEs when the initial plan encounters obstacles.

In this paper, we propose
\textbf{Re}asoning-P\textbf{a}th-E\textbf{di}ting (\readi), 
a novel framework that leverages the intrinsic planning ability of LLMs (Figure \ref{fig:framework}). In \readi, LLMs initially generate
a reasoning path, which is then instantiated on SEs to facilitate faithful reasoning. Path editing is triggered only if corrections are necessary.
This way, we alleviate the burden of step-by-step interaction for LLMs, resulting in improved overall efficiency. To harness the information of large-scale SEs, instead of injecting the entire static SEs into the model, we collect reasoning log as immediate feedback which includes details such as the position of stuck points, associated relations, half-way done instances, etc. This dynamic guidance refines the reasoning path more targeted and further enhances faithfulness. Our experiments on Question Answering over Knowledge Graphs (KGQA) and Tables (TableQA) demonstrate that \readi~surpasses existing solutions in terms of both LLM-calls and accuracy.

We summerize our contributions as follows:

$\bullet$ We introduce \readi, a novel framework where LLMs reason efficiently and faithfully over large-scale structured environments. Notably, \readi~is the first to fully harness LLMs intrinsic planning ability for reasoning in such contexts.

$\bullet$ In comprehensive experiments across five multi-hop reasoning tasks in KGQA and TableQA, \readi~outperforms other LLM-based solutions and surpasses most fine-tuned methods. Specifically, it achieves 67.0\% Hit@1 on CWQ, 78.7\% on WebQSP and \sota~results on MQA-1H. 

$\bullet$ We give detailed analysis which highlights the performance of \readi’s reasoning path generation and editing modules. Experiments demonstrate that \readi, with an average of 1.55 calls for editing, significantly reduces the number of LLM-calls compared to the step-by-step interaction paradigm (which costs 4 to 8 calls). Furthermore, the reasoning log reveals that \readi~exhibits characteristics akin to human thought process.

\section{Related Works}
\label{sec:related_work}

\noindent\textbf{Step-by-step reasoning over structured environments by LLMs.}
Introducing massive SEs (\textit{e.g.,}
Freebase~\cite{bollacker2008freebase} captures 45M entities and 3B facts over 100 domains)
into LLMs context windows is impractical.
Existing works break the task down to incrementally construct a reasoning path.
They either treat LLMs as an agent to invoke tools
based on history states
and observations ~\cite{liu2023agentbench, qin2023tool},
or design iterative procedures where LLMs are responsible for picking up items on SEs~\cite{gu2023dont,jiang2023structgpt,sun2023thinkongraph}. 
These works reach faithfulness by leveraging the reasoning ability of LLMs for \textit{tool or item selection}.
However, their performance is concerned with three shortcomings:
1) The iterative interaction with SEs is cumbersome, requiring quite a few LLM-calls, which is especially not efficient for complex reasoning tasks.
2) The greedy step-by-step decision lacks a global view of the path, making it prone to error propagation.
3) The accumulated prompts are lengthy where LLMs may lose attention of
history or candidates.
To ease these problems, we propose to directly generate a reasoning path, and edit it with feedback when the instantiation gets stuck.

\noindent\textbf{End-to-End reasoning over structured environments by fine-tuning.} Fine-tuned models \textit{memorize the environments} through training over annotations.
They either directly generate a path and then ground it on SEs~\cite{luo2023reasoning,huang2023question,shu2022tiara,hu-etal-2022-gmt},
or retrieve relevant items to build a path~\cite{Zhang_2022_sr,saxena-etal-2020-improving,ding2024enhancing}.
Such end-to-end reasoning shows efficiency with no interaction with SEs. However, they have three limitations:
1) The grounding of reasoning path only depends on model outputs, without ensuring faithfulness on SEs. To remedy this issue, they rely on a wider beam at the expense of more retrieved instances.
2) They rely heavily on annotations, which are expensive for massive environments.
3) The performance drops substantially on data unseen during training~\cite{Gu_grailqa,huang-etal-2023-markqa}, which is common in real-world scenarios.
To alleviate these problems, instead of fine-tuning, we propose to instantiate LLMs reasoning path. Then, if anything goes wrong, we call LLMs to edit the path. \looseness=-1

\textbf{Plan-and-Refine Reasoning with LLMs}. For faithfulness of LLMs reasoning, previous works adopt LLMs to refine the output \cite{pan2023automatically}. Some require LLMs to self-correct, prompting them to identify and correct errors~\cite{madaan2023selfrefine,pourreza2023dinsql}. Such methods achieve limited improvement, since they rely only on the intrinsic knowledge of LLMs, without any access to the environment.
Alternatively, other works require LLMs to refine the previous plan based on environmental feedback (\textit{e.g.} code error messages) \cite{chen2023selfdebug,taskweaver}. The feedback provides the execution results and possible errors, which are more purposeful and thus effective. However, how to collect feedback for large-scale structured environments remains an open question. In \readi, we collect immediate reasoning log through the instantiation of reasoning path, including the position of error, half-way done instances and associated relations.

\section{Task Definition}
\label{sec:preliminary}
\begin{figure}
    \setlength{\abovecaptionskip}{0.1cm}
    \setlength{\belowcaptionskip}{-0.5cm}
    \centering
    \includegraphics[scale=0.55]{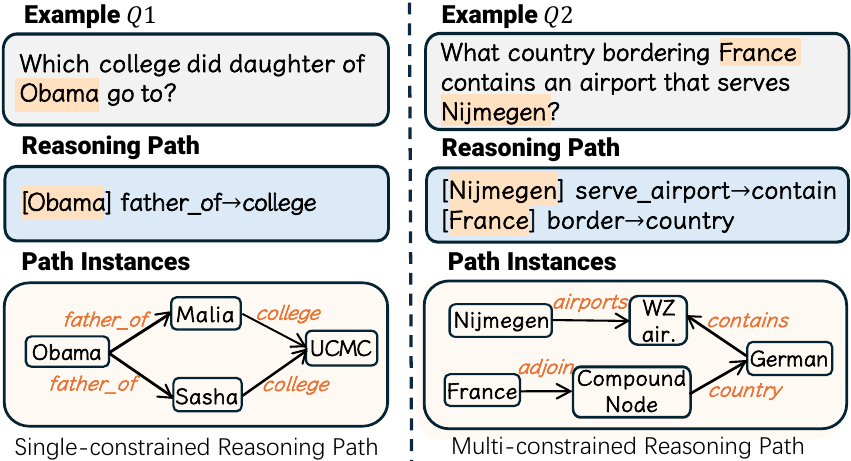}
    \caption{Examples of the question, reasoning path, and corresponding path instances on knowledge graph.}
    \label{fig:example_of_reasoning_path}
\end{figure}

Reasoning over structured environments (SEs), \textit{e.g.,} question answering over Knowledge Graphs (KGQA) and Tables (TableQA), typically requires matching a question with instances in SEs to constrain the answer.
An intermediate reasoning path is \textbf{a structural representation of the question}, as a bridge between the question and SEs.
Figure \ref{fig:example_of_reasoning_path} exemplifies some reasoning paths and their instances on KG. Note that a KG is a set of triple patterns, \textit{i.e., }$\{(e, r, e') | e, e' \in \mathcal{E}, r \in \mathcal{R}\}$, where $\mathcal{E}$ and $\mathcal{R}$ refers to the set of entity and relation, respectively.

Formally, given a question $Q$ and $n$ topic entities $E$,
the reasoning path $P$ is conditioned by several constraints in $Q$. It is worth noting that $P$ can represent complex constraints, (\textit{i.e.,} $P$ can be single-constrained or multi-constrained). 
A single-constrained $P$ is from only one topic entity. For example, a sequence of relations from the entity to the answer (Example $Q$1) or a Chain-of-thought~\cite{wei2023chainofthought}.
A multi-constrained $P$ is from multiple topic entities, consisting of multiple paths to constrain the answer (Example $Q$2). 
Correspondingly, we instantiate $P$ on KG to obtain instances. For Example $Q$1, the reasoning path \texttt{``father\_of$\rightarrow$university''} from \texttt{``Obama''} is instantiated to \texttt{``Obama$\xrightarrow[]{father\_of}$ Malia $\xrightarrow[]{college}$UCMC''} and \texttt{``Obama$\xrightarrow[]{father\_of}$ Sasha $\xrightarrow[]{college}$UCMC''}. 

\begin{figure*}
    \setlength{\belowcaptionskip}{-0.5cm}

    \centering
    \includegraphics[width=\textwidth]{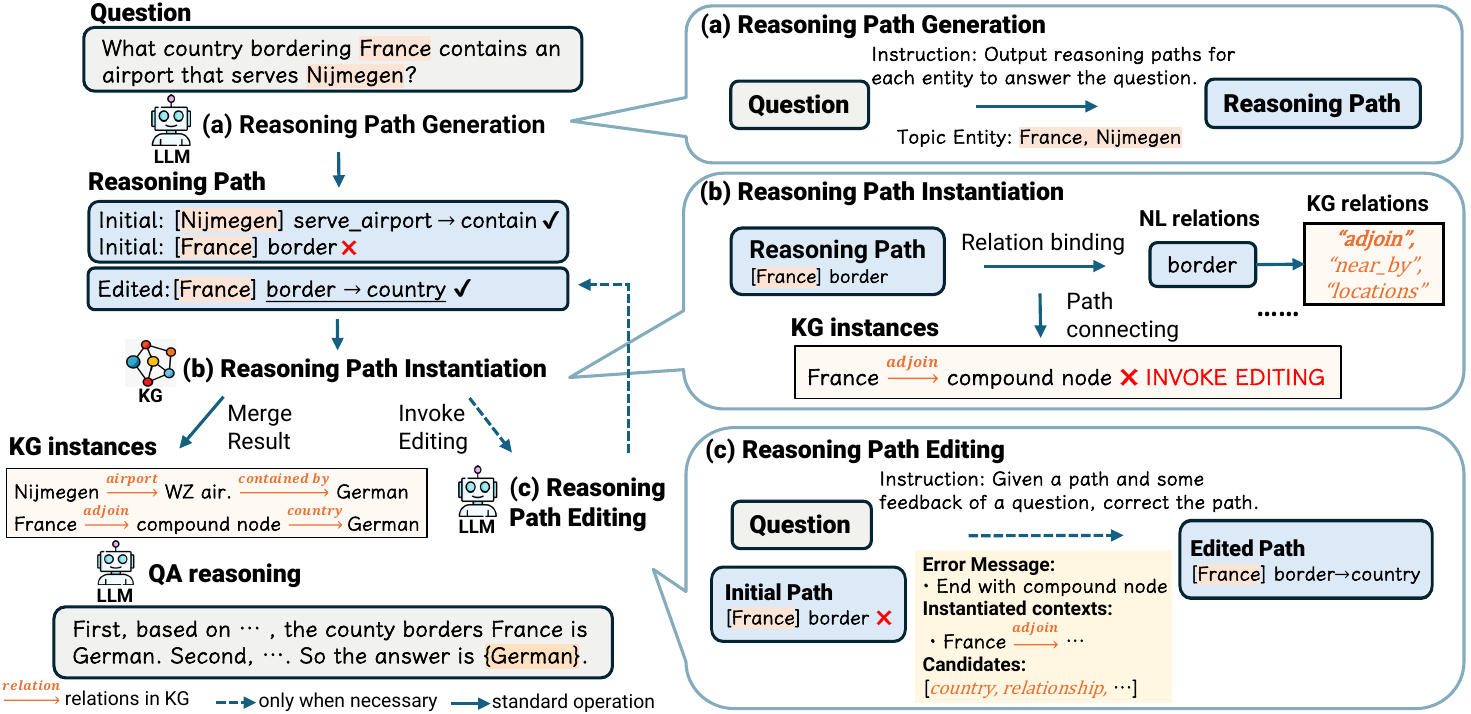}
    \caption{A running example of \readi~on KGQA. An LLM initially generates an reasoning path for a question. Then, we instantiate it on KG. If anything goes wrong (the path from \texttt{``France''}), we collect some error messages and call an LLM to edit the path. Finally, an LLM answers the question based on the KG instances.
    }
    \label{fig:running_example}
\end{figure*}

Following~\citet{sun-etal-2018graftnet},
we model reasoning over structured environments as an retrieve-then-reason task, where we leverage LLMs to build the reasoning path given $Q$. Then we reason over the path instances to obtain the answer.

\section{Methodology}
\label{sec:Methodology}

\subsection{Overview}
For better illustration, we adopt KGQA, a challenging scenario of multi-hop reasoning over massive environments, to showcase concrete implementation (Refer to Appendix \ref{sec:appendix_implementation_tableqa} for TableQA).
A running example is in Figure \ref{fig:running_example}.
Given a question and topic entities, we leverage the planning ability of LLMs to generate the initial reasoning path $P$ 
(Section \ref{subsec:reasoningpathgeneration}).
Then we instantiate $P$ on KG
(Section \ref{subsec:reasoninngpathinstantiation}).
If the entire $P$ is successfully instantiated, we take the intersection of all instances,
then generate the answer. If any path in $P$ gets stuck, we collect error messages to guide further refinement (Section \ref{subsec:Reasoning_Path_editing}).
Such process runs 
until $P$ is fully instantiated or a maximum edit time is reached. Please refer to Appendix \ref{sec:prompt_list} for concrete prompts of each module.

\subsection{Reasoning Path Generation}
\label{subsec:reasoningpathgeneration}

Inspired by~\citet{li2023fewshot}, we leverage in-context learning (ICL) to generate the initial reasoning path, as shown in Figure \ref{fig:running_example}(a).
Given a question and $n$ topic entities, we utilize Chain-of-Thought (CoT) to generate the initial path, consisting of $n$ constraints starting from each topic entity. For the example in Figure \ref{fig:running_example}, we have a two-constrained path (from \texttt{``France''} to constrain the bordering countries and from \texttt{``Nijmegen''} to airports serving it and then to countries containing these airports).

\subsection{Reasoning Path Instantiation}
\label{subsec:reasoninngpathinstantiation}
We instantiate the reasoning path $P$ on KG and merge the instances. The main difficulty lies in how to sequentially match the natural language (NL) relations in $P$ with relations in KG.
Our instantiation involves two steps: relation-binding and path-connecting, as shown in Figure \ref{fig:running_example}(b).

For relation-binding, given a path consisting of some NL relations (\textit{e.g.,} for
\texttt{``[France]~border''}, we have one relation \texttt{``border''} expressed in NL), we bind them to relation schemas in KG.
Since a relation $\boldsymbol{r}$ in NL may have similar relation schemas in KG with analogical format and semantic meanings \cite{li2023fewshot}, we leverage BM25 and Contriever~\cite{izacard2022unsupervised} to retrieve similar relations $\boldsymbol{\hat{r}}$ as candidates for $\boldsymbol{r}$.
For example, we bind \texttt{``border''} with 3 candidate relations ``\textit{adjoin, near\_by, locations}'' in KG (the orange italicized relations in Figure \ref{fig:running_example}(b).
This way, we obtain candidate relations
for all relations in $P$.

For path-connecting, note that the starting entity of each path is given. For the example in Figure \ref{fig:running_example}(b), we need to check if there exists any path instance in KG from \texttt{``France''} where relations in this instance sequentially match the bound candidate relations of all relations in $P$.
Specifically, if any relations in candidates 
``\textit{adjoin, near\_by, locations}'' connects to \texttt{``France''} 
(the orange bold ``\textit{adjoin}''), we obtain entities satisfying the constraint \texttt{``[France]~border''}, so the NL relation \texttt{``border''} is instantiated to ``\textit{\textbf{adjoin}}'' on KG. 
Then, we repeat the same process for the remaining relations in $P$ to finally instantiate the whole path.

If any path in $P$ is not successfully instantiated, \textit{e.g.,} none of the candidate relations $\boldsymbol{\hat{r}}$ connects to the current entity, this is the \textit{necessary time} to \textbf{invoke editing} (Section~\ref{subsec:Reasoning_Path_editing}).
If the entire $P$ is successfully instantiated or the maximum edit time is reached, we merge all KG instances by intersection to answer the question (Section \ref{subsec:qa_reasoning}). 
Please refer to Appendix \ref{appdx:reasoning_path_instantiation} for concrete implementation.

\subsection{Reasoning Path Editing}
\label{subsec:Reasoning_Path_editing}
The goal of editing is to help LLMs identify the error of previous reasoning path and provide some error messages, consisting of 2 steps: summarization and preparation, similar to an error traceback of coding, as shown in Figure \ref{fig:running_example}(c).

For summarization, we categorize the reasons of error as follows: \textit{i.}~``irrelevant NL relation $\boldsymbol{r_{err}}$'': none of the KG relation candidates of $\boldsymbol{r_{err}}$ connects to current entity; \textit{ii.}~``empty reasoning path''; \textit{iii}.~``path ends with compound nodes''\footnote{Compound value node, \textit{i.e.,} blank node, is typical in KG to express some complex entities, such as an event.}. Thus, if the instantiation goes wrong,
we detect error position $\boldsymbol{err}$, current NL relation $\boldsymbol{r_{err}}$ (none for reason \textit{iii}) and current ending entity $\boldsymbol{\hat{e}_{err}}$ (before which the NL relations are successfully instantiated). \looseness=-1

For preparation, we collect some useful messages: 1) the reason of error.
2) the currently halfway-done instances.
3) relations around $\boldsymbol{\hat{e}_{err}}$ is accessible, which might be the candidates to amend the halfway-done instances. Finally, an LLM is called to edit the previous path based on $Q$ and these error messages by ICL.

For the example in Figure \ref{fig:running_example}, we have the NL relation \texttt{``border''} instantiated to \textit{``adjoin''} and obtain a compound node at the end of instance, which triggers reason \textit{iii} to invoke LLMs for editing. 
Therefore, we concatenate reason \textit{iii}, halfway-done instance \texttt{``France$\stackrel{adjoin}{\longrightarrow}$Compound Node''} and candidate relations around the \texttt{``Compound Node''} (\textit{e.g.,} ``\textit{country}'' and ``\textit{relationship}'') as error messages. More details are in Appendix \ref{apdx:reasoning_path_editing}. 

\subsection{QA Reasoning}
\label{subsec:qa_reasoning}
Upon obtaining the merged KG instances $\mathcal{S}_q$, \textit{i.e.,} the intersection of instances for each path (constraint), we build an
LLM-based reasoning module to answer the question $Q$.
We concatenate $Q$ and $\mathcal{S}_q$ as input and prompt an LLM to generate the answer. Note that the form of $\mathcal{S}_q$ is a set of triple patterns $(entity, relation, entity')$. For entities, we convert the mid in KG to the corresponding friendly name. For relations and compound nodes, we keep their original forms in KG.
Specifically, we ask LLMs to pick the knowledge triples used to answer $Q$ in a CoT manner.
For the example KG instances from \texttt{``France''} in Figure \ref{fig:running_example}, we have 2 knowledge triples \texttt{``(France, adjoin, compound node), (compound node, country, German)''}, based on which, the LLM can reason that \texttt{``the country bordering France''} is \texttt{``German''}.

\section{Experiments}
\label{sec:Experiments}
To comprehensively evaluate the reasoning ability of \readi~on large-scale structured environments, we experiment on two complex tasks, KGQA and TableQA.
(Please refer to Appendix \ref{sec:appendix_implementation_tableqa} for detailed implementation of TableQA).

\subsection{Datasets}
\label{sec:dataset}
We evaluate on three KGQA and two TableQA datasets. Detailed statistics are in Appendix \ref{sec:dataset_statistics}. 

\noindent \textbf{WebQuestionsSP (WebQSP)}~\cite{yih-etal-2016-value} contains KGQA questions from Google query logs with up to 2-hop reasoning on Freebase, mostly requiring a single-constrained reasoning path.

\noindent \textbf{ComplexWebQuestions (CWQ)}~\cite{Talmor2018TheWA} is a complex KGQA benchmark, challenging for up to 4-hop reasoning on Freebase, with 55\% multi-constrained questions.

\noindent \textbf{MetaQA}~\cite{zhang2017variational} is a KGQA dataset from movie domain,
with 3 levels of difficulty, denoted as MQA-1H, MQA-2H and MQA-3H.

\noindent \textbf{WikiTableQuestions~(WTQ)}~\cite{pasupat2015compositional} contains questions over 421 tables, challenging for complex aggregation operations, \textit{e.g.,} count, argmax, and sorting.

\noindent \textbf{WikiSQL}~\cite{zhong2017seq2sql} is a large-scale complex dataset based on Wikipedia tables, requiring comparison, aggregation and arithmetic operations.


\begin{table*}[ht]
    \setlength{\abovecaptionskip}{0.2cm}
    \setlength{\belowcaptionskip}{-0.5cm}
    
    \centering    
    \begin{tabular}{l cc  ccc}
    \toprule
     Methods & WebQSP & CWQ & MQA-1H &MQA-2H &MQA-3H \\  \midrule
     
     \multicolumn{6}{c}{\textit{Training-based Method}} \\ 
     EmbedKGQA~\cite{saxena-etal-2020-improving} & 66.6 & - & \textbf{97.5} & {98.8} & 94.8 \\
     NSM~\cite{He_2021} & 67.7 & 47.6 & \underline{97.1} & $\underline{99.9}$ & {98.9} \\
     TransferNet~\cite{shi2021transfernet} & 71.4 & 48.6 & 97.5 & $\textbf{100}^{\star}$ & $\textbf{100}^{\star}$ \\
     SR+NSM+E2E~\cite{Zhang_2022_sr} &69.5 & 49.3 &  - & - & - \\ 
     UniKGQA~\cite{jiang2023unikgqa} &75.1 & 50.7 & \textbf{97.5} & {99.0} & \underline{99.1}\\ 
    ReasoningLM~\cite{jiang2023reasoninglm} & \underline{78.5} & $\textbf{69.0}^{\star}$ &96.5 &98.3 & 92.7\\ 
        RoG~\cite{luo2023reasoning} & $\textbf{85.7}^{\star}$ & \underline{62.6} & - & - & 84.8  \\ \midrule[1.5pt]%
        
      \multicolumn{6}{c}{\textit{Inference-based Method}} \\ 

    Davinci-003~\cite{ouyang2022training} & 48.7 & - & 52.1 & 25.3 & 42.5\\
     GPT3.5 \cite{gpt35} & 65.7 & 44.7 & 61.9 & 31.0 & 43.2\\  
     GPT4~\cite{openai2023gpt4} & 70.7 & 52.1 & 71.8 &52.5 & 49.2 \\  
    AgentBench~\cite{liu2023agentbench} & 47.8 & 24.8 & - & - & - \\  
     StructGPT~\cite{jiang2023structgpt} & 69.6 & - & 97.1 & \underline{97.3} & 87.0\\

    \midrule
 \readi-GPT3.5  & \underline{74.3} & \underline{55.6} & $\underline{98.4}$ & $\textbf{99.9}$ & $\textbf{99.4}$   \\
   \readi-GPT4 &\textbf{78.7}& \textbf{67.0} & $\textbf{98.5}^{\star}$ & $\textbf{99.9}$ & \underline{99.2} \\

     \bottomrule
    \end{tabular}%
    \caption{QA performance (Hit@1) of \readi~on KGQA datasets. Results of GPT3.5, GPT4~\cite{openai2023gpt4} and AgentBench~\cite{liu2023agentbench} are run by ourselves, others are from the origin paper.
    Bold and underline fonts denotes the best and second-best for two types of methods, respectively. $\star$ denotes the overall \sota~result.
    }
    \label{tab:main_results_KGQA}
\end{table*}

\subsection{Baselines}
\label{baseline_methods}
\textbf{KGQA Baselines.}
\textit{Training-based methods} fine-tune pre-trained language models (PLMs):
EmbedKGQA~\cite{saxena-etal-2020-improving}, 
NSM~\cite{He_2021}, 
TransferNet~\cite{shi2021transfernet},
SR+NSM+E2E~\cite{Zhang_2022_sr},
UniKGQA~\cite{jiang2023unikgqa},
ReasoningLM~\cite{jiang2023reasoninglm} and RoG~\cite{luo2023reasoning}.
\textit{Inference-based methods} call LLM-APIs:
Davinci-003~\cite{ouyang2022training}, GPT3.5, GPT4~\cite{openai2023gpt4}, AgentBench~\cite{liu2023agentbench}, StructGPT~\cite{jiang2023structgpt}.
All baselines assume golden topic entities are given. 

\noindent \textbf{TableQA Baselines.}
\textit{Training-based methods} find-tune PLMs: TAPAS~\cite{Herzig_2020}, UnifiedSKG~\cite{xie2022unifiedskg}, TAPEX~\cite{liu2022tapex}.
\textit{Inference-based methods} call LLM-APIs: Davinci-003~\cite{ouyang2022training}, GPT3.5, GPT4~\cite{openai2023gpt4}, StructGPT~\cite{jiang2023structgpt}. Note that \textit{Inference-based methods} model TableQA as an information retrieval task.

Following \citet{tan2023chatgpt}, we adopt Hit@1, assessing whether the predicted entity is correct, to evaluate KGQA. We adopt denotation accuracy, assessing whether the prediction exactly matches the golden, to evaluate TableQA.
We elaborate baselines in Appendix \ref{sec:baselines}. We also discuss in-depth comparison with ToG \cite{sun2023thinkongraph} and DATER \cite{DATER_ye} in Appendix~\ref{comparison_with_tog}.

\subsection{Implementation Details}
We adopt \texttt{gpt-3.5-turbo} \cite{gpt35} and GPT4~\cite{openai2023gpt4} as LLMs, denoted as \readi-GPT3.5 and \readi-GPT4.  
Temperature is 0.3 for all modules.
For relation-binding, we deploy a Pyserini as a hybrid searcher with BM25 and Contriever~\cite{izacard2022unsupervised}.
For each relation generated by LLMs, we retrieve top 5 similar relations on Freebase. 
For instantiation, we deploy a Virtuoso server following the instructions\footnote{https://github.com/dki-lab/Freebase-Setup}. More details can be found in Appendix \ref{apdx:detailed_setup}.

\subsection{Main Results}
\textbf{Results for KGQA} As shown in Table \ref{tab:main_results_KGQA}, 
overall, \readi~significantly outperforms all Inference-based methods, the vanilla LLMs and most training-based methods on all datasets, demonstrating the effectiveness of \readi.
Compared with inference-based methods, \readi~substantially boosts the results of the vanilla LLM (by 8.6\% on WebQSP and 14.9\% on CWQ), demonstrating that \readi~enables LLMs to practically interact with structured environments.
Moreover, \readi-GPT3.5 already significantly surpasses \sota~results with LLM-APIs (by 4.7\% on WebQSP and 12.4\% on MQA-3H), and \readi-GPT4 further enhances the performance.
With fewer LLM-calls, our directly-generated reasoning path and editing framework still achieves better performance.
Compared with Training-based methods, without large-scale supervision and cost of beam search, \readi~achieves comparable performance (\textit{e.g.} 67.0 Hit@1 on CWQ) by some demonstration examples, showing the effectiveness of \readi. 
Additionally, \readi~sets new \sota~results MQA-1H, showing the effectiveness of editing which provides pertinent feedback upon instantiation errors.

 \begin{table}[!tb] 
    \setlength{\abovecaptionskip}{0.2cm}
    \setlength{\belowcaptionskip}{-0.65cm}
    \centering    
    \begin{tabular}{lcc}
    \toprule
     Methods & WTQ &WikiSQL\\ \midrule
      \multicolumn{3}{c}{\textit{Training-based Method}} \\ 

     TAPAS &48.8 &83.6\\
     UnifiedSKG~(T5-3B) &49.3 &86.0\\
     TAPEX &\textbf{57.5} &\textbf{89.5}\\
\midrule[1pt]
      \multicolumn{3}{c}{\textit{Inference-based Method}} \\ 
     Davinci-003 & 34.8 & 49.1 \\
     GPT3.5 & 55.8 & 59.8 \\
     GPT4 & 57.0 & 59.9 \\
    StructGPT & 52.2 & 65.6\\
    \midrule
     \readi-GPT3.5  & \textbf{61.7} & \textbf{66.2} \\
     \readi-GPT4  & 61.3 &  66.0 \\

     \bottomrule
    \end{tabular}
    \caption{QA performance (denotation accuracy) of \readi~on TableQA datasets.
    Bold fonts denotes the best results for Training-based and Inference-based methods.}
    \label{tab:tableqa_results}
\end{table}

 \begin{table*}[!tb] 
     \setlength{\abovecaptionskip}{0.1cm}
     \setlength{\belowcaptionskip}{-0.4cm}

    \centering    
    \begin{tabular}{lcccccccc}
    \toprule
 \multirow{2}*{Variance of \readi} & \multicolumn{4}{c}{Answer Coverage Rate (AC)} & \multicolumn{4}{c}{QA Performance (Hit@1)} \\
\cmidrule{2-9} 
&  \textit{Corrupt} & \textit{Empty} & GPT3.5 & GPT4  &  \textit{Corrupt} & \textit{Empty} & GPT3.5 & GPT4 \\  \midrule
    w/o edit  & - & - & 56.7 & 62.7  & - & - & 51.0 & 57.2 \\
    w/ edit by GPT3.5&54.0 & 56.4  & 62.5  & 64.3  & 57.3 & 58.5 & 58.7  & 58.5 \\ 
    w/ edit by GPT4 &55.6 &63.9& 68.6 & 65.8 &58.2 & 59.9 & 58.1 & 59.3 \\ 
    
     \bottomrule
    \end{tabular}%
    \caption{Answer Coverage Rate (AC) and QA Performance (Hit@1) of variance of \readi~(GPT3.5 as reasoning module). Each column denotes a path generation method. \textit{Corrupt} means a path with some randomly-sampled relations. \textit{Empty} means empty path. w/o edit means we only leverage the initial reasoning path.\looseness=-1
    }
    \label{tab:ablation_study}
\end{table*}

\noindent\textbf{Results for TableQA}  We experiment on TableQA scenario requiring multi-hop reasoning over tables to show the generalizability of \readi. As shown in Table \ref{tab:tableqa_results},
overall, \readi~outperforms all Inference-based and most Training-based baselines, setting \sota~results on WTQ, demonstrating the effectiveness of our framework.
Compared with Inference-based methods, \readi~surpasses previous \sota~methods by 9.5\% and the vanilla LLM by 5.9\% on WTQ, showing that \readi~significantly improve the LLMs performance.
Compared with Training-based methods, \readi, without massive annotations, is significantly superior on WTQ, while trailing behind on WikiSQL. This may be due to the i.i.d. distribution between the training and testing sets of WikiSQL, favoring the results of fully-trained methods.
Interestingly, \readi-GPT3.5 are comparable with \readi-GPT4. This might because we asks LLMs to reason (with aggregation operations) directly based on retrieved table items.
Further analysis shows that \readi-GPT3.5 has more chance of invoking editing than \readi-GPT4, which offers more pertinent feedback from the environments. Elaboration on effectiveness of editing is in Section \ref{sec:ablation_study}, Section \ref{reasoningpathanalysis} and Appendix \ref{performance_edit_time}.

\section{Analysis}
\label{sec:analysis}
We further analyze \readi's modules, reasoning path and efficiency on 1000 test samples of CWQ. For fairness, we base all evaluation on our instantiation and reasoning modules. Please refer to Appendix \ref{apdx:detailed_ana} for more detailed analysis of \readi.

\subsection{Ablation Study}
\label{sec:ablation_study}
\textbf{Effectiveness of reasoning path generation and editing modules.}
As shown in Table~\ref{tab:ablation_study}, we adpot the answer coverage rate (AC, rate of instances containing the answer) and QA performance (Hit@1) for illustration. We also analyze the \textbf{robustness of editing module} with a \textit{Corrupt} path
and an \textit{Empty} path.
First, \readi~establishes a plug-and-play nature
for both modules, showing their effectiveness. With only an initial path (we instantiate the path and maintain the longest path if it goes wrong), \readi~reaches comparable results (1st row). With an \textit{Empty} initial path (\textit{Empty} columns), the editing modules still yield competitive results close to the full \readi.
Second, editing with LLMs significantly improves the performance (2nd and 3rd rows), further showing the effectiveness of the editing module.
Third, generally, higher capacity of LLMs leads to better results for both modules, which meets our expectation.
Lastly, \readi~shows robustness for reasoning path editing, performing well even with an \textit{Empty} or even a \textit{Corrupt} path (\textit{Corrupt} and \textit{Empty} columns).

\subsection{Reasoning Path Analysis}
\label{reasoningpathanalysis}
We compare
reasoning path of \readi~with representative fine-tuned methods, \textit{i.e.,} 
Subgraph Retrieval (SR)~\cite{Zhang_2022_sr} trains an encoder to retrieve relation and then builds a path, RoG~\cite{luo2023reasoning} tunes a Llama 2~\cite{touvron2023llama} to generate a path.

 \begin{table}[!tb] 
    \setlength{\abovecaptionskip}{0.1cm}
    \setlength{\belowcaptionskip}{-0.55cm}
    \centering    
    \begin{tabular}{lccc}
    \toprule
  \multirow{2}*{Methods} & \multicolumn{2}{c}{Graph Quality}  & \multicolumn{1}{c}{QA Perf.} \\
  \cmidrule{2-4}
     & AC & \#RK & Hit@1 \\  \midrule
     SR &  &  & \\
      \hspace{0.5em} - beam size 1 & 58.4 & \textbf{26.3} & 50.9 \\
      \hspace{0.5em} - beam size 3  & 67.2 & 47.1 & 54.6 \\ 
    RoG  &  &  & \\ 
      \hspace{0.5em} - beam size 1 & 57.0 & 69.5 & 52.2  \\
      \hspace{0.5em} - beam size 3  & \textbf{77.5}& 170.1 & 57.3 \\ 
      
    \midrule
     \readi~initial path  &   & \\
      \hspace{0.5em} - GPT3.5& 56.7 & 134.6  & 51.0\\
      \hspace{0.5em} - GPT4 & 62.7 & 101.4 & 57.2  \\
    \readi~full & & \\
      \hspace{0.5em} - GPT3.5& 62.5 & 93.7  & 58.7 \\
    \hspace{0.5em} - GPT4& 71.8 & 121.5  & \textbf{59.3} \\
     \bottomrule
    \end{tabular}
    \caption{Reasoning path evaluation of \readi and compared methods. AC and \#RK denotes answer coverage rate and number of retrieved knowledge, respectively.\looseness=-1
    }
    \label{tab:init_plan_compare}
\end{table}

\noindent\textbf{Quality of \readi's reasoning path.} We adopt answer coverage rate (AC)
and number of retrieved knowledge (\#RK) as the quality of graph. Ideally, the higher AC and lower \#RK, the better. Also, we analyze the QA performance (Hit@1), shown in Table~\ref{tab:init_plan_compare}.
First, \readi's initial path is comparable with fine-tuned methods, with GPT4 surpassing them by a large margin (5.0\% and 6.3\% Hit@1 than RoG and SR, respectively), showing the effectiveness of our reasoning path.
Second, with some necessary editing, \readi~obtains substantially higher AC, with a little higher \#RK (GPT4) and even lower \#RK (GPT3.5), and significantly higher QA performance than fine-tuned methods, showing the effectiveness of LLMs editing. 
Third, with wider beam size, fine-tuned methods gain higher AC, yet a drastically growing \#RK, and lower Hit@1 than \readi, illustrating our superiority.

\noindent \textbf{Extensive features of \readi's reasoning path.} The quality of reasoning path is multi-dimensional.
To further show insights of reasoning path by LLMs,
we design some metrics, including the instantiation progress, error types, etc.
We compare with SR, RoG and Golden (Outermost in Figure \ref{fig:relative_compare_quality}). More detailed analysis is in Appendix \ref{appdx:detailed_features_of_reasoning_path}.

\begin{figure}
    \setlength{\abovecaptionskip}{0.1cm}
    \setlength{\belowcaptionskip}{-0.6cm}
    \centering
    \includegraphics[scale=0.5]{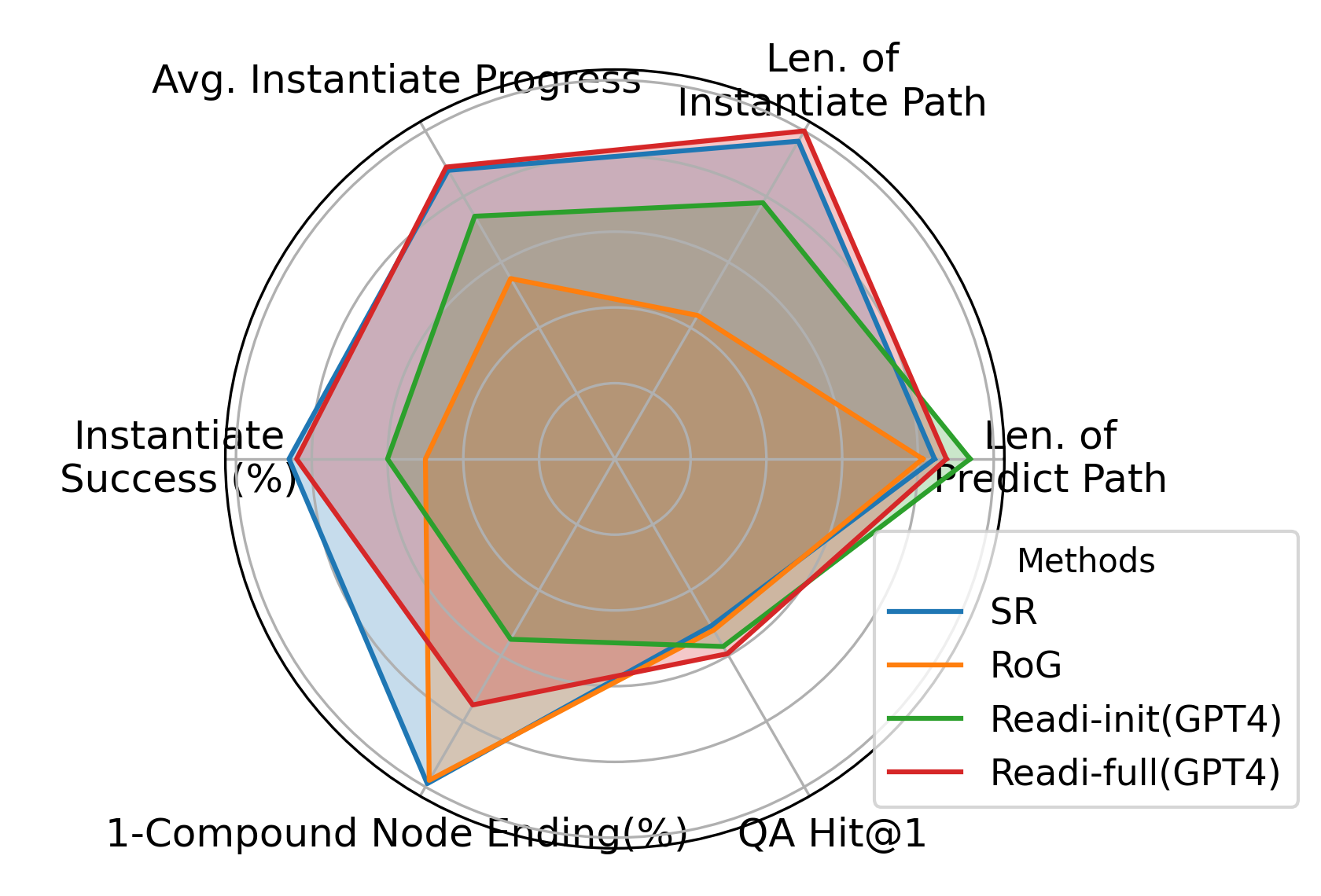}
    \caption{Extensive features of \readi's reasoning path, compared with fine-tuned methods and Golden.}
    \label{fig:relative_compare_quality}
\end{figure}

Insights driven by Figure \ref{fig:relative_compare_quality} are three-folded: 
1) \readi's initial path already achieves better in QA results.
More than a half (60\%) of \readi's initial paths can be fully instantiated (LLMs called only once), showing that \readi~effectively unleash the strong planning ability of LLMs.
For each path, a large proportion (averagely 74\%) can be instantiated, again indicating the possibility to safe LLM-calls. 
2) With necessary editing, \readi~gets significantly closer to golden (From green to red in Figure \ref{fig:relative_compare_quality}), exceeding compared methods in many metrics, especially the QA results. 
3) Interestingly, the path by LLMs establish a human-like nature, for humans tend to get stuck at points they have never seen in real world. For Compound Node Ending Rate (special for KG), fine-tuned methods are close to Golden, showing that they memorize the structured information. However, they trail behind behind \readi~on QA results, demonstrating that 
{unfaithfulness} still exits for fine-tuned methods. \looseness=-1

\begin{table}[bt] 
    \setlength{\abovecaptionskip}{0.1cm}
    \setlength{\belowcaptionskip}{-0.3cm}
    \centering
    \small
    \begin{tabular}{p{7.5 cm}}
    \toprule
    \textbf{Question 1: } What is the name of the money used in the country the Peruvian Paso breed originated?\\
    \textbf{Initial Reasoning Path:} (from ``Peruvian Paso'')\\
biology.organism.breeds$\rightarrow$biology.breed.originated\_in$\rightarrow$ location.country.currency\_used\\
          \textbf{Error Message:} irrelevant relation ``biology...breeds''.\\ Candidates: biology.breed.originated\_in, ... \\
    \textbf{Edited Reasoning Path:} (from ``Peruvian Paso'')\\
    biology.breed.originated\_in$\rightarrow$location.country.currency\_used\\
    \midrule
    \textbf{Question 2: } What to see in the country that has Gozo?\\
    \textbf{Initial Reasoning Path:} (from ``Gozo'')\\
    location.location.containedby$\rightarrow$location.country.attractions\\
        \textbf{Error Message:}~irrelevant relation ``location...attractions''.\\
        Contexts: Gozo $\xrightarrow[]{location.location.containedby}$ Melta \\
        Candidates: travel.travel\_destination.tourist\_attractions, ... \\
        \textbf{Edited Reasoning Path:} (from ``Gozo'')\\
location.location.containedby$\rightarrow$travel.travel\_destination.tour-ist\_attractions \\
    \bottomrule
    \end{tabular}
    \caption{Cases of \readi's reasoning path editing.
    }
    \label{tab:case_study}
\end{table}

\subsection{Efficiency Evaluation}
\begin{figure}
    \setlength{\abovecaptionskip}{-0.1cm}
    \setlength{\belowcaptionskip}{-0.3cm}
    \centering
    \includegraphics[scale=0.62]{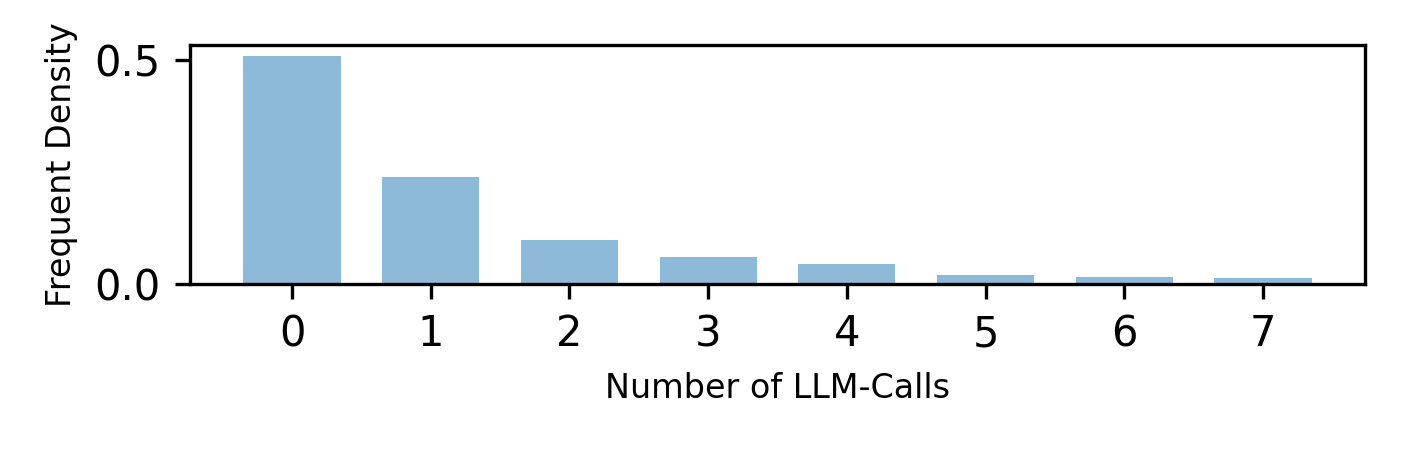}
    \caption{Distribution of number of LLM-Call for reasoning path editing of \readi-GPT4.}
    \label{fig:efficient_distribution}
\end{figure}

\textbf{How many LLM-calls do we need?} Due to unexpectable nature of LLMs output, we cannot give an exact number of LLM-calls for \readi. Instead, we present the distribution of number of LLM-calls for editing in Figure \ref{fig:efficient_distribution} and some exemplar cases in Table \ref{tab:case_study}.
Note that there are at least 2-hop and up to 4-hop reasoning required for CWQ (theoretically 4-8 LLM-calls for iterative interaction). The instantiation success rate and average instantiate progress in Figure \ref{fig:relative_compare_quality} already demonstrate that \readi~can save a bunch of LLM-calls. In Figure \ref{fig:efficient_distribution}, most of the time the initial reasoning path is already instantiable, no need for more LLM-calls. Averagely, the LLM is called 1.55 times to edit the path, saving more invocation than iterative interaction. Also, cases in Table \ref{tab:case_study} show that, with the necessary editing, LLMs can correct previous path based on some error messages during instantiation.

\label{subsec:efficiency_eval}

\section{Conclusion}
\label{sec:con_and_future}

In this paper, we propose a novel framework \readi~where LLMs can reason over structured environments efficiently and faithfully. 
In \readi, LLMs initially generate a reasoning path which is then instantiated on environments to facilitate faithful reasoning. Path editing is triggered only if the instantiation gets stuck. We showcase the implementation of \readi~on knowledge graph, evaluate the effectiveness on KGQA and TableQA and analyze extensive features of \readi. Our work also shed lights on practically interaction between natural language and structured environments, where LLMs plays a crucial role to bridge the gap.

\section{Limitations}
\label{sec:limitation}
Although \readi~achieves strong performance in complex reasoning task over structured environments, \textit{e.g.,} knowledge graph and tables, there are still some limitations of our method. First, we leverage only two LLMs as backbone to evaluate \readi. Therefore, more experiments can be done to test performance of \readi~for other LLMs (including the inference-based and training-based). Second, we adopt reasoning path as a representation of natural language utterance and model the task as a information-retrieval one. An interesting direction is semantic parsing, \textit{e.g., } text2SQL, which we will leave for future work. Moreover, we can also test \readi~on question answering over database or other reasoning fields. Third, the instantiation in our current implementation of KGQA is intuitive but may induce many queries for SPARQL engine if encountered some ``big'' entities.
Finally, our proposal, to fully harness intrinsic planning ability of LLMs and incorporate structural reasoning log to edit the path, performs well and saves some LLM-calls in experiments, which does not mean we should always depend on LLMs plans. And although the initial path by LLMs can be well instantiated, this just means the path by LLMs corresponds to some instances on KG, but not ensures that the path exactly matches with the constraints in the question.

\section{Acknowledgements}

The authors would like to thank Jue Zhang, Yuzhong Qu, Dapeng Li, Jinmao Li, Kaikai An, Yurong Wu and Wentao Ding for their valuable feedback. The authors would also like to thank Xiaokang Zhang for providing the reasoning path of Subgraph Retrieval.

\bibliography{custom}

\appendix

\clearpage

\section{Implementation of \readi~for TableQA}
\label{sec:appendix_implementation_tableqa}

\begin{table*}[ht]
    \centering    
    \begin{tabular}{ccccc}
    \toprule
     Name in English & Name in Turkish & Area~($km^2$) & Depth & Location (Districts and/or provinces) \\  \midrule
        Lake Van & Van Gölü & 3755 $km^2$ & 171 m & Van, Bitlis\\ 
     Lake Tuz & Tuz Gölü & 1500 $km^2$ & 2 m & Aksaray, Ankara, Konya\\
     Lake Palas Tuzla & Palas Tuzla Gölü & 106 $km^2$ & 15 m&Palas/Kayseri\\
     \multicolumn{5}{c}{\textbf{......}}\\
     \bottomrule
    \end{tabular}
    \caption{An example from WTQ \cite{pasupat2015compositional}, with the question \texttt{``which is deeper, lake tuz or lake palas tuzla?''}.
    }
    \label{tab:example_of_tableqa}
    
\end{table*}
Algorithm \ref{algo:readi} demonstrates \readi's framework.
Our implementation of \readi~for TableQA strictly sticks to Algorithm \ref{algo:readi}, and is simpler than KGQA. The main difference between TableQA and KGQA is the implementation of reasoning path (Refer to Section \ref{sec:preliminary}). Detailed prompts of all modules is in Appendix \ref{sec:prompt_list}.

\begin{algorithm}
\footnotesize
\caption{\readi}
    \label{algo:readi}
    \KwData{task $Q$, entity set $E$, environment $\mathcal{G}$, max edit time $T$}
    $t \leftarrow 0$ \\
     $P_0 \leftarrow \textbf{Reasoning\_Path\_Generation}(Q, E)$ \label{algo:init_plan}\\
    \While{$t \leq T$}{ 
         $I_{\mathcal{G}}, Err_{msg} \leftarrow \textbf{Instantiate}(P_t, \mathcal{G})$
         \label{algo:intanstiation} \\
        \eIf{$ Err_{msg} \neq \emptyset $}{
          $P_{t+1}\leftarrow\textbf{Edit\_Path}(Err_{msg}, P_t, Q)$ \label{algo:plan_edit}\\
         $t \gets t+1$\\}{
            \textbf{Go to Line 11}}
        }
     $Instances_{\mathcal{G}} \leftarrow \textbf{Merge\_Results}(I_{\mathcal{G}})$ \label{algo:merge_result}
\end{algorithm}

\textbf{Reasoning path} for TableQA starts from a given table, to some columns and then to some rows, in order to constrain specific items in the table to answer the question. A sample of WTQ \cite{pasupat2015compositional} dataset is shown in Table \ref{tab:example_of_tableqa}. The reasoning path of such question is from this table to column \texttt{``Name in English''} and \texttt{``Depth''}, and then to row items whose \texttt{``Name in English''} is \texttt{``Lake Tuz''} and \texttt{``Lake Palas Tuzla''}.

For \textbf{reasoning path generation}, we ask LLMs to generate a reasoning path, given a question and some table descriptions (\textit{i.e.,} the header and a randomly-sampled row), by in-context learning. Specifically, we require the LLMs to pick up at least to rows for complex arithmetic and aggregation operations in TableQA. For the example in Table \ref{tab:example_of_tableqa}, ideally, the LLM would generate a dict indicating the chosen header is \textit{``['Name in English', 'Depth']''}, and the row constrain is \textit{``{"Name in English": ['Lake Tuz', 'Lake Palas Tuzla']}''}.

For \textbf{reasoning path instantiation}, since schemas of a table is not as massive as a KG, we don't need a relation binding module to retrieve candidate relations and connect them according to the reasoning path. 
Instead, we first filter out columns and then filter out the rows in the reasoning path. For the example in Table \ref{tab:example_of_tableqa}, we filter out the column \texttt{"Name in English"} and \texttt{"Depth"} and rows whose \texttt{"Name in English"} is \texttt{"Lake Tuz"} or \texttt{"Lake Palas Tuzla"}.
If the the instantiation goes wrong, this is the \textit{necessary time} to invoke editing. Here we only consider whether the instantiation of columns in the reasoning path goes wrong. If a selected row in the reasoning path fails to be instantiated, we do not invoke editing and return all rows in the columns. 

For \textbf{reasoning path editing}, we collect the currently instantiated columns and candidate columns when the instantiation goes wrong, as error messages. Note that we only consider wrong columns as error, where we categorize the reasons of error as follows: \textit{i.}~``irrelevant column $\boldsymbol{col_{err}}$'': the column $\boldsymbol{col_{err}}$ in the reasoning path failed to be instantiated in the given table; \textit{ii.}~``insufficient columns in reasoning path'': the output reasoning path contains less than two columns (most questions in WTQ \cite{pasupat2015compositional} and WikiSQL \cite{zhong2017seq2sql} need at least two columns to constrain the answer).
Then, we ask the LLMs to edit previous reasoning path according to the feedback. For example, if chosen header is \textit{["English Name"]}, not matching the headers in Table \ref{tab:example_of_tableqa}. We provide all candidate headers and a randomly-sampled row in Table \ref{tab:example_of_tableqa} as error messages.

The \textbf{QA reasoning} is similar to KGQA, we concatenate the question and instantiated table items and ask the LLMs to answer the question.

\section{Detailed Implementation of \readi~for KGQA}
\label{sec:appendix_implementation_KGQA}

\subsection{Reasoning Path Instantiation Details}
\label{appdx:reasoning_path_instantiation}
In this part, we demonstrate more details of 
path-connecting mentioned in Section \ref{subsec:reasoninngpathinstantiation}.
Assume that we have a reasoning path starting from an entity, with $m$ relations $R$. All relations are bound to candidates $\hat{R}$.
For the example in Figure \ref{fig:running_example}, we have a reasoning path \texttt{``[Nijmegen] serve\_airport $\rightarrow$ contain''} and have \texttt{``$r_1:$~serve\_airport''} bound to ``$\hat{r}_1:$~\textit{airport, terminal, serving port}'' in KG, and similarly have \texttt{``$r_2:~$contain''} bound to ``$\hat{r}_2:~$\textit{contained by, contains, place of}''. The purpose if path-connecting is to check if there exists an instance in KG where the starting point is entity \texttt{``Nijmegen''}, then connect to some relations in $\hat{r}_1$, and then connect to some relations in $\hat{r}_2$.

The path-connecting is essentially an Breadth-first search (BFS) algorithm, with a queue $que$ containing possible entities connected to current relations. Firstly, we push the starting point (\textit{e.g.,} \texttt{``Nijmegen''}) into $que$. Second, at each time, we pop out current head of $que$ and check if current candidate relations (\textit{e.g.,} ``\textit{adjoin, terminal, serving port}'') can connect to it on KG. If so, we successfully instantiate \texttt{``serve\_airport''} on KG, and we can obtain some entities which are \texttt{``serve\_airport''} of \texttt{``Nijmegen''} (\textit{e.g.,} \texttt{``WZ air.''} and \texttt{``NTA.''}). Then, we add these entities to $que$. Such BFS searching continues until all relations are instantiated or anything goes wrong.

Note that for a ``big'' entity, their may be a substantial number of tail entities for a relation. For example, there are hundreds of entities for \texttt{``France$\stackrel{location}{\longrightarrow}$''}. This may end up with a longer instantiation time and shed lights on future improvements of \readi. In this case, we adopt a threshold to constrain the size of $que$.

\subsection{Reasoning Path Editing Details}
\label{apdx:reasoning_path_editing}
Section \ref{subsec:Reasoning_Path_editing} covers the basic process of reasoning path editing. Here are some implementation details for the preparation step, the purpose is to adopt pertinent structural information on KG to help LLMs identify the error position of previous reasoning path and edit it. For the half-way done instances in the Error Messages, we use ``compound node'' to indicate all cvt nodes in Freebase ~\cite{bollacker2008freebase}, and we just sample several (as a hyper-parameter) instances of each instantiated relation to showcase the path instances in KG of the previous path. The loss of information from sampling is minor because the messages just inform the LLMs that the relations in the reasoning path can be grounded to certain instances on KG. For candidate relations, we also adopt a threshold to constrain the size, if there are too many of them, we filter out using similarity search (same embedding as relation-binding) according to the original question.

For example, there are hundreds of compound node instances connected to \texttt{``France$\stackrel{adjoin}{\longrightarrow}$''}, and we just use \texttt{{``France$\stackrel{adjoin}{\longrightarrow}$compound node''}} to represent the instances. If their are hundreds of candidate relations connected to these compounds nodes, we use the question \texttt{``What country bordering France contains an airport that serves Nijmegen''} to filter out top 35 similar candidates.

\section{Detailed Experimental Setups}
\label{apdx:detailed_setup}
\subsection{Dataset Statistics}
\label{sec:dataset_statistics}
We experiment \readi~on test sets of all datasets. We adopt Hit@1 and denotation accuracy for KGQA and TableQA, respectively. We evaluate \readi~on 3 KGQA and 2 TableQA datasets (Descriptions in Section \ref{sec:dataset}). 
Statistics of datasets are shown in Table \ref{tab:dataset_statics}. Note that we model all datasets as an \textit{information retrieval} task, instead of a \textit{semantic parsing} one.






\begin{table}[t]
\centering
    \begin{tabular}{cccc}
    \toprule
    \textbf{Dataset} & \textbf{Training}  & \textbf{Dev} & \textbf{Test}  \\
    \midrule  
    \multicolumn{4}{c}{\textit{KGQA Dataset}} \\
    \textsc{WebQSP} & 3,098  &  -   &  1,639 \\
    \textsc{CWQ} & 27,639  & 3,519  & 3,531 \\
    \textsc{MQA-1H} & 96,106 &  9,992 &  9,947 \\
    \textsc{MQA-2H} & 118,980 &  14,872 &  14,872 \\
    \textsc{MQA-3H} & 114,196 &  14,274 &  14,274 \\ \midrule
    \multicolumn{4}{c}{\textit{TableQA Dataset}} \\
    \textsc{WTQ} & 11,321  &  2,831   &  4,344 \\
    \textsc{WikiSQL} & 56,355  &  8,421   &  15,878 \\
     \bottomrule   
    \end{tabular} 
\caption{Statistics of experiment datasets.
} 
\label{tab:dataset_statics}
\end{table}

\subsection{LLM API version}

For CWQ and WebQSP, we use gpt-3.5-turbo-16k-0613 and gpt-4-32k for GPT3.5 and GPT4, respectively. For others, we adopt gpt-3.5-turbo-0613 and gpt-4 for GPT3.5 and GPT4, respectively.

\subsection{Baselines}
\label{sec:baselines}
\textbf{KGQA baselines.} 
\textit{Training-based methods:}
\begin{itemize}
    \item 
 EmbedKGQA~\cite{saxena-etal-2020-improving} adopts an encoder to retrieve relevant entities and generate the answer.
    \item 
 NSM~\cite{He_2021} adapts neural state machine for KGQA and rank entities in a retrieved subgraph.

    \item 
\noindent TransferNet~\cite{shi2021transfernet} adopts a transparent framework to rank entities according to different parts of a question in a subgraph.

    \item 
\noindent SR+NSM+E2E~\cite{Zhang_2022_sr} trains an encoder to retrieve relevant relations and build a path from retrieved relations.

    \item 
\noindent UniKGQA~\cite{jiang2023unikgqa} retrieves a sub-graph and rank the schemas in a unified way.

    \item 
\noindent ReasoningLM~\cite{jiang2023reasoninglm} designs an entity encoding and training framework to rank entities in a sub-graph.

    \item 
\noindent RoG~\cite{luo2023reasoning} trains a LLama 2 \cite{touvron2023llama} to firstly generate a path, second ground this path to the knowledge graph, and then generate answer based on the grounded graph.
\end{itemize}
\textit{Inference-based methods:}

\begin{itemize}
    \item 
\noindent 
Davinci-003~\cite{ouyang2022training}, GPT3.5 and GPT4~\cite{openai2023gpt4} are based on LLM-APIs. We adopt few-shot in-context learning to ask the model to output the answer of question in order to test the models inherent knowledge of datasets.
    \item 
\noindent AgentBench~\cite{liu2023agentbench} is an agent-based method asking LLMs to call tools based on tool-discription, history and observations.

    \item 
\noindent StructGPT~\cite{jiang2023structgpt} requires LLMs to iterative pick up relations and entities based on current returned candidates.
\end{itemize}

\textbf{TableQA baselines.}
\textit{Training-based methods:} 
\begin{itemize}
    \item TAPAS~\cite{Herzig_2020} predicts denotation by selecting table cells and optionally applys an aggregation operator to the selection.
    \item UnifiedSKG~\cite{xie2022unifiedskg} sequentializes the table and tunes a T5-3B model to answer the question.
    \item TAPEX~\cite{liu2022tapex} guides language models to mimic a SQL executor.
\end{itemize}

\textit{Inference-based methods:} 
\begin{itemize}
    \item Davinci-003~\cite{ouyang2022training}, GPT3.5, GPT4~\cite{openai2023gpt4} are based on LLM-APIs. We adopt few-shot in-context learning to ask the model to output the answer, based on the question and the entire table.
    
    \item StructGPT~\cite{jiang2023structgpt} iteratively filter out columns and rows of table and ask the LLMs to answer the question.
\end{itemize}
Note that all Inference-based baselines are information retrieval ones.

\section{Detailed Analysis}
\label{apdx:detailed_ana}
\subsection{Elaboration on Reasoning Path Analysis}
\label{appdx:detailed_features_of_reasoning_path}
Note that quality of reasoning path is multi-dimensional.
To further show some insights of reasoning path by LLMs,
we meticulously design some metrics.
We introduce average Length of predicted path (LPP), average length of instantiated path (LIP), average instantiation progress (AIP, average of LIP/LPP for each question), instantiation success rate (ISR) and Compound nodes ending rate (CER), where all length mentioned above refers to number of relations on path. 
We compare with fine-tuned SR~\cite{Zhang_2022_sr} and RoG~\cite{luo2023reasoning} (beam=1). 
The Golden Path is obtained by extracting relations from golden logical forms in the dataset, which is only used for analysis.

\begin{table}[!tb] 
    \centering    
    \begin{tabular}{@{\hspace{0.6em}}l@{\hspace{0.3em}}c@{\hspace{0.3em}}c@{\hspace{0.3em}}c@{\hspace{0.3em}}c@{\hspace{0.3em}}c}
    \toprule
     Metrics & LPP & LIP & AIP & ISR & CER \\  \midrule
     Golden Path & 3.2 & 3.2 & 1.0 & 1.0 &0\\ \midrule
    SR~\cite{Zhang_2022_sr}& 3.7 &3.3&0.88 &0.86 & 0.01\\ 
     RoG~\cite{luo2023reasoning}&2.6 & 1.4&0.55 &0.50 &0.02\\ \midrule

     \readi-GPT3.5 - init  & 4.3 & 2.5 & 0.64 & 0.46 & 0.49 \\
     \readi-GPT3.5 - full &3.3 & 2.8& 0.86 &0.80 &0.22\\
     \readi-GPT4 - init   & 3.4 & 2.5 &0.74 &0.60&0.45\\
     \readi-GPT3.5 - full & 3.6 & 3.2 &0.89 &0.84&0.25 \\
     \bottomrule
    \end{tabular}%
    \caption{Extensive features of \readi-GPT3.5 and \readi-GPT4's reasoning path, compared with Golden reasoning path. LPP, LIP, AIP, ISR, IER means length of predict path, length of instantiated path, average instantiation progress, instantiation success rate and ending with intermediate nodes rate, respectively.
    }
    \label{tab:init_path_metrics}
    
\end{table}

\begin{figure}
    \setlength{\abovecaptionskip}{0.1cm}
    \setlength{\belowcaptionskip}{-0.3cm}

    \centering
    \includegraphics[scale=0.62]{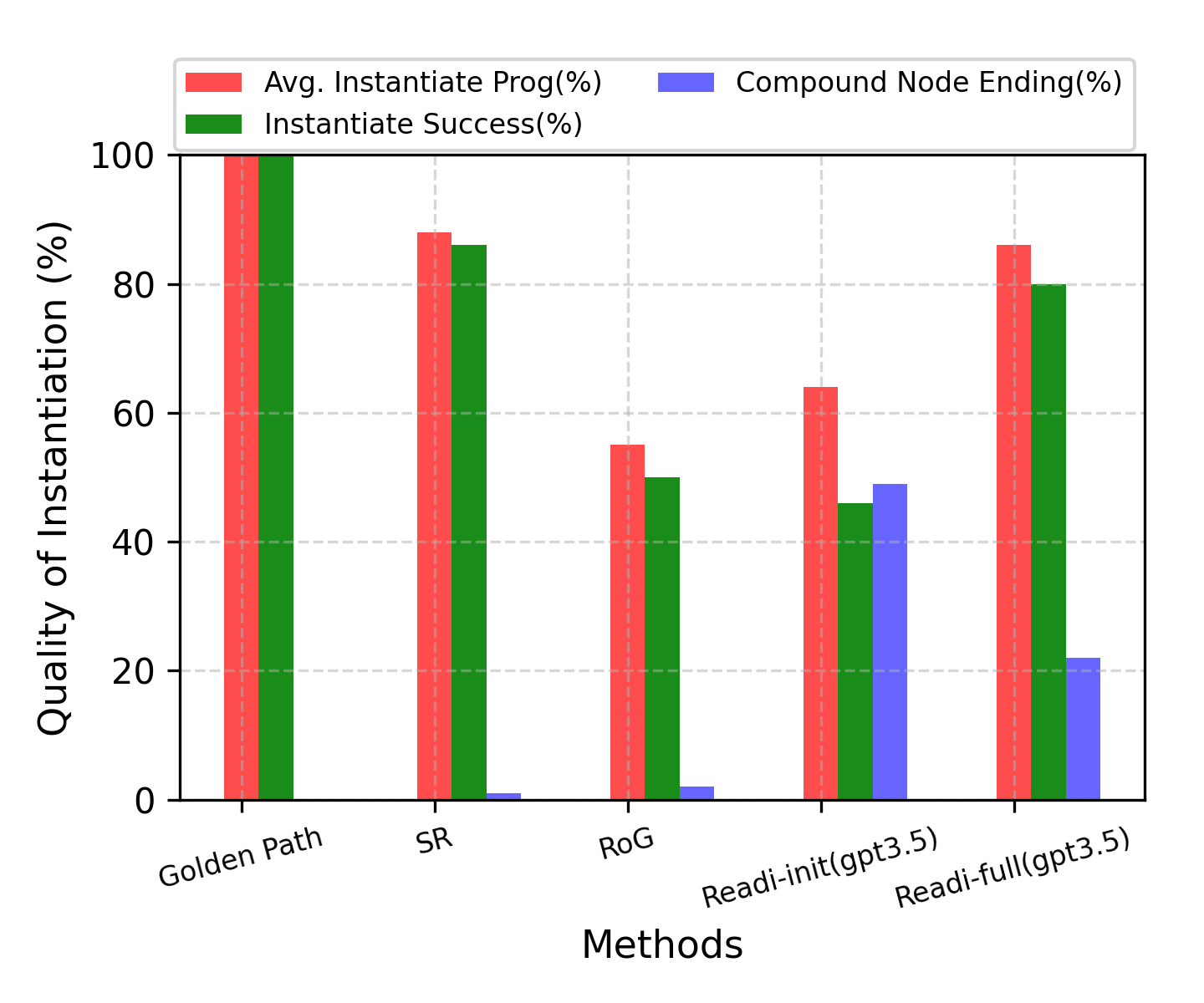}
    \caption{Extensive analysis of \readi-GPT3.5 reasoning path, compared with Golden reasoning path.}
    \label{fig:exact_quality_of_path}
\end{figure}

\begin{figure}
    \setlength{\belowcaptionskip}{-0.45cm}
    \centering
    \includegraphics[scale=0.5]{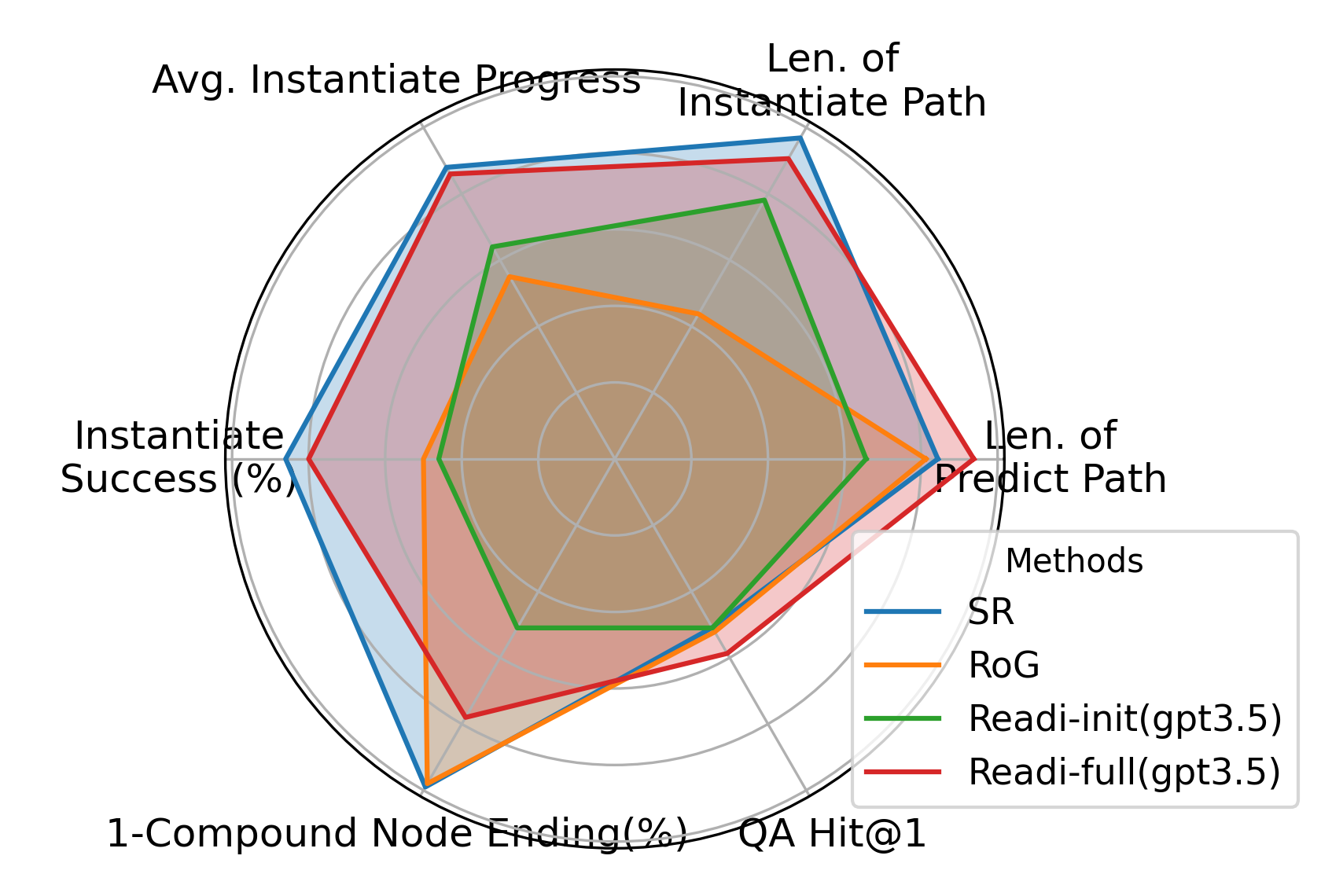}
    \caption{Features of \readi's reasoning path, compared with fine-tuned methods and Golden.}
    \label{fig:relative_compare_quality_35}
\end{figure}

\begin{figure}
    \setlength{\abovecaptionskip}{-0.1cm}
    \centering
    \includegraphics[scale=0.60]{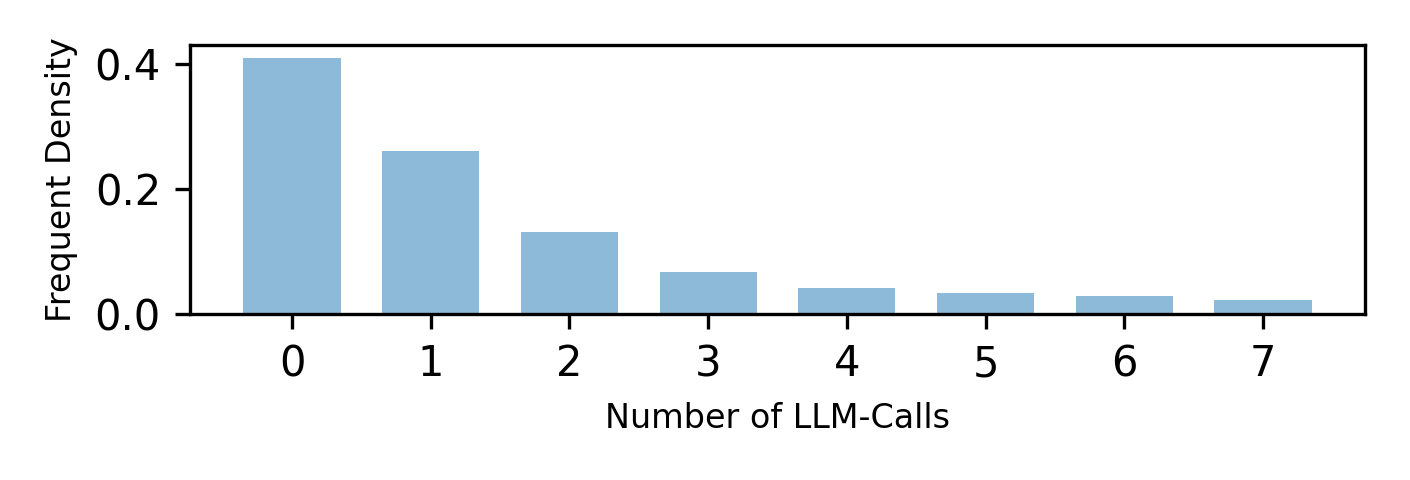}
    \caption{Distribution of number of LLM-Call for reasoning path editing of \readi-GPT3.5}
    \label{fig:efficient_distribution_35}
\end{figure}

The absolute performance of different metrics of \readi~are shown in Table \ref{tab:init_path_metrics}. Also, Figure \ref{fig:exact_quality_of_path} demonstrates some instantiate progress of \readi-GPT3.5. Figure \ref{fig:relative_compare_quality_35} shows some relative results of different metrics for \readi-GPT3.5, compared with fine-tuned methods and the Golden. Additionally, we show the distribution of number of LLM-calls for editing for \readi-GPT3.5 in Figure \ref{fig:efficient_distribution_35}.

It is shown that the initial reasoning path by \readi~is already comparable with fine-tuned ones. With some necessary editing, \readi~significantly gets closer to golden. In addition, without awareness of the environments, a large number of \readi's initial path (49\%) get stuck at compound nodes, while fine-tuned methods well memorize the schemas in structured environments. However, with pertinent feedback upon instantiation errors, \readi~substantially alleviates this problems and reaches higher QA performance than fine-tuned methods, again demonstrating that unfaithfulness still exits even with large-scale fine-tuning.

\subsection{Performance of varied MAX\_EDIT\_TIME}
\label{performance_edit_time}
We have reported the distribution of Editing times in Figure \ref{fig:efficient_distribution} (GPT4) and Figure \ref{fig:efficient_distribution_35} (GPT3.5) to show the efficiency of \readi. To some extent, the distribution also showcases the quality of the Reasoning Path and Effectiveness of Editing module. Note that we set the MAX\_EDIT\_TIME to eight to test the quality, which can be set smaller to save more resources. To further provide some insights, we experiment on different MAX\_EDIT\_TIME setting on CWQ subset in Table \ref{tab:max_edit_time_exp}. Results show that with more chances to edit, the reasoning path gets better, further showing the effectiveness of Editing. And four times of editing already achieves an comparable outcome.

\begin{table}[!tb] 
    \centering    
    \begin{tabular}{lcccc}
    \toprule
    MAX\_EDIT\_TIME	& 2	&4 &6 & 8\\  \midrule
Hit@1 &	51.6&	59.2&	59.7&	59.8\\
     \bottomrule
    \end{tabular}%
    \caption{\readi-GPT35's performance for different MAX\_EDIT\_TIME on CWQ.}
    \label{tab:max_edit_time_exp}

\end{table}

\subsection{Performance of Retriever}
The performance of retrieval may affect the instantiation. We calculate the recall of retriever in Table \ref{tab:perf_retriever}, showing that the performance of retrival is not the bottleneck of \readi. Moreover, our implementation is adapted from KB-BINDER \cite{li2023fewshot}, which has already shown effectiveness of binding relations to KG.

\begin{table}[!tb] 
    \centering    
    \begin{tabular}{lcc}
    \toprule
   & GPT35	&GPT4\\  \midrule
   CWQ&	84.6	&86.4\\
WebQSP	&95.5	&91.9\\
     \bottomrule
    \end{tabular}%
    \caption{Recall (\%) of retriever for instantiation.}
    \label{tab:perf_retriever}

\end{table}

\begin{table*}[bt] 
    \centering    
    \begin{tabular}{lcccc}
    \toprule
 Error Types & Instantiation Error	&Answer not Covered	&QA Error	&False Negative\\  \midrule

CWQ-GPT3.5	&22\%&	28\%	&30\%	&20\%\\
WebQSP-GPT3.5&	6\%&	32\%&	32\%&	30\%\\
CWQ-GPT4	&18\%&	36\%	&20\%&	26\%\\
WebQSP-GPT4	&20\%	&26\%	&22\%	&32\%\\
     \bottomrule
    \end{tabular}%
    \caption{Qualitative error analysis of \readi.}
    \label{tab:error_analysis}

\end{table*}

\begin{table}[!tb] 
    \centering    
    \begin{tabular}{lcc}
    \toprule
   & CWQ	&WebQSP\\  \midrule
avg token (k)	&52.9	&37.6\\
avg cost (\$)	&0.090&	0.064\\
     \bottomrule
    \end{tabular}%
    \caption{Average token cost of Readi (reasoning path generation and editing.}
    \label{tab:token_cost}

\end{table}

\subsection{Token Cost}

As indicated in our title, we discusses how LLMs can efficiently and faithfully interact with structured environments. By efficiency, we focus on \textit{the way of interaction}, i.e. less LLM invocations (Section \ref{intro}), to obtain pertinent information on large-scale structured environments for multi-hop reasoning. As in some industrial circumstances, the times of LLM-calls are strictly constrained. However, some might concern the token cost of \readi~as a measure of efficiency, though it is not our priority. Here, we report the average token cost per question in Table \ref{tab:token_cost}. We utilize Python library \textit{tiktoken}, for GPT3.5 with six-shot for generation, 5-shot for editing and MAX\_EDIT\_TIME set to 4 for CWQ and 2 for WebQSP. 

Note that it is non-trivial to directly compare the \textit{exact token cost} with other methods, for the sake of number and format of examples, uncontrollable LLM-API traffic, and even the difficulty of the task. For example, StructGPT \cite{jiang2023structgpt} claims to incorporate 32-shot examples, but the exact few-shot prompts are not provided. \readi's token cost can also be modulated by adjusting the MAX\_EDIT\_TIME.
Furthermore, excessive calls or overly challenging tasks may cause the LLM-API to fail in providing the appropriate content or format, exceeding the MAX\_TRY\_TIMES, which also affects the token cost. That's also why we analyze the times of LLM-call as efficiency of our interaction framework in the main body of paper.

\subsection{Qualitative Error Analysis}

We analyze 200 randomly-sampled error cases of CWQ and WebQSP in Table \ref{tab:error_analysis}. We divide the error into four categories (Section \ref{subsec:Reasoning_Path_editing} and \ref{reasoningpathanalysis}): \textit{i.} "Instantiation Error": Even after Editing, the path is still not fully instantiated.
\textit{ii.} "Answer not Covered": Answer not in retrieved knowledge. \textit{iii.} "QA Error": Even the answer is covered, the QA output is still wrong (e.g., hallucination).
\textit{iv.} "False Negative": For example, the ground truth of \texttt{``the two continents Turkey is in''} is \texttt{``Eurasia''}, and the model output \texttt{``Europe and Asia''}.

Based on Table \ref{tab:error_analysis}, first, our reasoning path can be well instantiated and utilized to answer the multi-hop reasoning question. Second, there is still room for improvement of the retrieved knowledge, which we would like to focus in our future work. Third, the hallucination in QA reasoning still exits, with GPT4 performing better than GPT3.5.

\section{Generalizability}
\label{apdx:generalizability}
Note that our focus is not on comparative analysis of various LLMs. We propose \readi~to enhance LLMs reasoning over structured environments. Therefore, we utilize LLMs with strong understanding and reasoning capabilities (\textit{e.g.,} GPT3.5) to show their performance with \readi. Here we discuss generalizability of \readi~framework.

\noindent \textbf{Generalizability to domain-specific KG.} The concept of reasoning path is a structured representation of a multi-hop reasoning process (Section \ref{sec:preliminary}). By the \textit{intrinsic planning ability} in Section \ref{intro}, we adopt LLMs strong question understanding and reasoning ability to directly generate the reasoning path. Here is an intuition of \readi:
we humans can navigate multi-hop reasoning challenges across various domains by recognizing named entities and relations of the task. LLMs, trained on large-scale natural language corpus, may develop these abilities. Therefore, \readi~is designed to operate without the need of domain-specific knowledge.

We have experimented on both large-scaled KG (CWQ and WebQSP) and domain-specific KG (MetaQA), where \readi~shows impressive performance (Section \ref{sec:Experiments}). One concern is generalization to specific domain with massive relations. Here, we illustrate how \readi~handles such cases with an example.

Consider a question \texttt{``What's the nationality of Lebron James''}. Assume that the LLM has few domain knowledge and initially generates \texttt{``[Lebron James] nationality''} as the reasoning path, whereas the ground truth path in KG is \texttt{``[Lebron James] born\_in - city\_of''}.

When instantiation (Section \ref{subsec:reasoninngpathinstantiation}), we bind \texttt{``nationality''} to relations in the KG but find none of them connected to \texttt{[Lebron James]}, so we \textbf{invoke editing}. In the error message, we include relations around \texttt{[Lebron James]}, which is \texttt{[born\_in, sex, father\_of, …]}, as a hint for Editing (Section 4.4 and Appendix B.2).

With the semantic understanding that \texttt{``born\_in''} can relate to \texttt{"nationality of Lebron James"}, the LLM can correct previous path. The ablation of Editing (Section \ref{sec:ablation_study}) and analysis of Reasoning Path after Editing (Section \ref{reasoningpathanalysis}) further demonstrate these. Note that relations around \texttt{[Lebron James]} is limited, compared with those in the whole KG.

The whole idea mirrors human cognitive strategies when conducting web searching for multi-hop reasoning tasks, where we flexibly adjust our sub-conscious plan based on information in browsers.

\begin{table}[!tb] 
    \centering    
    \begin{tabular}{lcc}
    \toprule
        LLM	& CWQ	& WebQSP \\  \midrule

    Llama-2-70b-chat	&49.2	&76.9 \\
     \bottomrule
    \end{tabular}%
    \caption{\readi's performance with other LLM-API on CWQ and WebQSP.}
    \label{tab:generalizability_llama2}
    
\end{table}

\noindent \textbf{Genalizability to other LLMs.}
Here, we show \readi's geralizability to other LLMs. Due to API availability, we test \readi~with Llama-2-70b-chat on CWQ and WebQSP samples. The MAX\_EDIT\_TIME is set to 4 for CWQ and 2 for WebQSP. For fair comparison, we adopt our GPT3.5-based QA reasoning module. Table \ref{tab:generalizability_llama2} shows that \readi~can generalize to other open-sourced LLMs.

We also discuss \readi's generalizability to fine-tuned models.
RoG (Section \ref{reasoningpathanalysis}) tunes a Llama 2 to generate a reasoning path, showing the feasibility of generating a path by fine-tuning. We've shown that \readi~performs well with Editing by LLM to improve the path. One concern is about Editing with a fine-tuned model, depending on annotations, which we would go deeper in future works.

\section{Comparison with ToG and DATER}
\label{comparison_with_tog}

We further compare \readi~with ToG \cite{sun2023thinkongraph} and DATER \cite{DATER_ye}.

\subsection{With ToG}
Think-on-Graph (ToG) is an LLM-based KGQA method, which also adopts an LLM to reason the answer in an information retrieval manner. Here, we discuss the differences between our \readi~and ToG. As discussed in Related Works (Section \ref{sec:related_work}), ToG adopts an LLM to step-by-step filter out some Knowledge Graph (KG) instances, similar to beam search. Note that there is no reasoning path, \textit{i.e., a structural representation of the question by LLMs}, in ToG's methodology. 

\noindent \textbf{Interaction Paradigm.} 
Our \readi's novelty lies in the following aspects: \textbf{1) Required Capability.} In ToG, the LLM traverses on KG, similar to beam search, to filter out entities or relations by \textit{scoring all candidates at each step}, which mainly requires discrimination ability. In contrast, our
\readi~requires the strong understanding and reasoning ability to \textit{maintain a structural representation} of the entire question.
\textbf{2) Way to Introduce the Environment.} To obtain structural information in large-scale KG, ToG introduces all relations around an entity, or all tail entities around a relation, as candidates for LLMs to \textit{score the distribution}. Such process can be cumbersome because there are sometimes massive candidates on KG, \textit{e.g.,}~\texttt{locations in France}. On the other hand, in \readi, we collect pertinent reasoning log only upon the instantiation errors. \readi~fully unleash the understanding ability of LLMs and ease the burden of multi-turn scoring for long candidate lists.
\textbf{3) Grounding.} To obtain instances, in ToG, LLMs select items at each step, restricted by the beam size, hindering it from reasoning questions requiring logical operations (based on a set of instances for aggregation, comparison, etc). Conversely, for \readi, we introduce a novel instantiation module (Section \ref{subsec:reasoninngpathinstantiation}) to obtain all instances based on constraints in the reasoning path.

\noindent \textbf{Generalizability.} \readi~can generalize to more circumstances (Appendix \ref{apdx:generalizability}): \textbf{1) Supported Environments}, ToG focuses on KG. However, \readi~is a more general interaction framework for reasoning over structured environments. We showcase the concrete implementation of both KGQA (Section \ref{sec:Methodology}) and TableQA (Appendix \ref{sec:appendix_implementation_tableqa}). \textbf{2) Supported Reasoning Tasks.} ToG, whose KG instances are restricted by the beam size, falls short for questions requiring logical operations, \textit{e.g.,} aggregation, comparison. However, \readi, adopting LLMs to maintain a structural representation of the question, is not restricted by beam size, which can cover such reasoning cases by LLMs reasoning with retrieved instances.

\noindent \textbf{Experimental Results.} We do not cover the results of ToG in main results, for the sake of fair comparison. The evaluation metric from the official published code of ToG differs from the standard Hit@1. Conversely, \readi's evaluation strictly follows \citet{tan2023chatgpt}, which is also adopted by all compared LLM-based baselines. While ToG's reported results are not reproducible, here we report \readi's results based on ToG's metrics in Table \ref{tab:tog_compare}, where our \readi, with less LLM-calls, still outperforms ToG, overall.

\begin{table}[!tb] 
    \centering    
    \begin{tabular}{lcc}
    \toprule
        Method	&  CWQ	& WebQSP \\  \midrule

    ToG-GPT3.5	& 57.1 &76.2 \\
    \readi-GPT3.5	& \textbf{57.9}	& \textbf{77.5} \\ \midrule
    ToG-GPT4	& 67.6 &\textbf{82.6} \\
    \readi-GPT4	& \textbf{69.2}	& 82.4 \\
     \bottomrule
    \end{tabular}%
    \caption{\readi's performance compared with ToG over ToG's metric.}
    \label{tab:tog_compare}
    
\end{table}

\subsection{With DATER}

We have considered DATER but did not directly compare with it for the following reasons.

\noindent \textbf{The base LLMs.}
DATER adopts Codex \cite{codex} as the backbone LLM, which is different from ours. Codex performs significantly better than GPT3.5 in TableQA (9.1\% higher acc on WTQ \cite{liu2023rethinking}). However, Codex is close-soured and does not offer API-calls any more\footnote{https://platform.openai.com/docs/deprecations}. Therefore, it is not fair to directly compare \readi~(based on GPT3.5) with DATER.

\noindent \textbf{The way of reasoning over Tables.}
Our experiment is to show if our interaction framework, \readi, can adopt LLMs to effectively obtain useful information from structured environments and then reason the answer. Therefore, we compare with methods modeling TableQA as an \textit{Information Retrieval} task (refer to Section \ref{sec:preliminary} and Section \ref{baseline_methods}). However, DATER uses a text2SQL model to obtain facts in table. This involves external tools (the SQL Interpreter) to handle the logical and numerical operations (e.g., aggregation, compare and calculation), which is not fair to compare with.

\section{Prompt List}
\label{sec:prompt_list}

Note that we do not modify prompts for baseline methods. Prompts for the vanilla LLMs is in Table \ref{tab:vanilla_prompt}, following \cite{sun2023thinkongraph}. For TableQA, the prompts for vanilla LLMs is the same in Table \ref{tab:prompt_list_table}. 
Detailed prompts for each module of \readi~is in Table \ref{tab:prompt_list} (KGQA) and Table \ref{tab:prompt_list_table} (TableQA).

For number of few-shot demonstrations, on CWQ \cite{Talmor2018TheWA} and WebQSP \cite{yih-etal-2016-value} we adopt 6 shots for reasoning path generation and 5 shots for other modules. For MQA\cite{zhang2017variational}, the shot number for reasoning path generation follows \cite{li2023fewshot} and we adopt only 3 shots for editing and reasoning (we donnot design a reasoning module for MQA). For WTQ \cite{pasupat2015compositional} and WikiSQL \cite{zhong2017seq2sql}, we use 7, 2, 7 shots for generation, editing and reasoning, respectively.
\begin{table*} 
    \setlength{\abovecaptionskip}{0.1cm}
    \setlength{\belowcaptionskip}{-0.2cm}
    \centering

    \begin{tabular}{p{15.5 cm}}
    \toprule
    \textbf{Instruction} Please answer the question:\\
    \textbf{Demonstration Example}\\
Q: What state is home to the university that is represented in sports by George Washington Colonials men's basketball?\\
A: {Washington, D.C.}.\\
\bottomrule
    \end{tabular}
    \caption{Prompts for vanilla LLMs for KGQA in our experiments.}
    \label{tab:vanilla_prompt}
\end{table*}

\begin{table*} 
    \setlength{\abovecaptionskip}{0.1cm}
    \setlength{\belowcaptionskip}{-0.5cm}
    \centering

    \begin{tabular}{p{15.5 cm}}
    \toprule
    \textbf{Prompts for reasoning path generation} \\Given a question and some Topic Entities in the Question, output possible freebase Relation Paths starting from each Topic Entities in order to answer the question. \\
    \textbf{Demonstration Example}\\
Question: Find the person who said ``Taste cannot be controlled by law'', where did this person die from? \\
Topic Entities: [``Taste cannot be controlled by law''] \\
Thought: There is only one topic entity, the answer is constrained by one path. 
For, the path from ``Taste cannot be controlled by law'', firstly, it should cover the person quote it. Second, it should cover the place where the person died. \\
Path: $\{$
``Taste cannot be controlled by law'': $[$
    Taste cannot be controlled by law $\rightarrow$ people.person.quotations $\rightarrow$ people.deceased\_person.place\_of\_death 
$]\}$\\
\midrule
    \textbf{Prompts for reasoning path editing} \\Task: Given an Inital Path and some feedback information of a Question, please correct the initial path.\\
    \textbf{Demonstration Example}\\
Question: The movie featured Miley Cyrus and was produced by Tobin Armbrust?\\
Initial Path: Miley Cyrus$\rightarrow$film.film.actor$\rightarrow$film.film.producer
Error Message\\
1. <compound node> in the end.\\
2. relation "film.film.producer" not instantiated.\\
Instantiation Context\\
Instantiate Paths: Miley Cyrus $\rightarrow$ film.actor.film $\rightarrow$ <compound node> \\
Candidate Relations\\$[$'film.director.film', 'film.performance.film', ...$]$\\
Corrected Path\\
Goal: The Initial Path starts from Miley Cyrus, which should cover the movies featured by Miley Cyrus.\\
Thought: In Instantiate Paths I know that Miley Cyrus acts some films, described by a compound node.
In candidates, I find "film.performance.film" most relevant to get the films.
Meanwhile, "film.film.producer" is not relevant to my Goal.\\
Final Path: Miley Cyrus$\rightarrow$film.actor.film$\rightarrow$ film.performance.film\\
\midrule
    \textbf{Prompts for QA reasoning} \\
    Given a question and the associated retrieved knowledge graph triplets (entity, relation, entity), you are asked to answer the question with these triplets. If the given knowledge triples is not enough or missing, you can use your own knowledge. Use \{\} to enclose the answer! Please think step by step.\\
    \textbf{Demonstration Example}\\
Q: The artist nominated for The Long Winter lived where? \\
Knowledge Triplets:\\(The Long Winter, book.written\_work.author, Laura Ingalls Wilder),
(Laura Ingalls Wilder, people.person.places\_lived, m.28e5697),
(m.28e5697, people.place\_lived.location, De Smet)\\
A: First, based on (The Long Winter, book.written\_work.author, Laura Ingalls Wilder), the author of The Long Winter is Laura Ingalls Wilder. Second, based on (Laura Ingalls Wilder, people.person.places\_lived, m.28e5697), (m.28e5697, people.place\_lived.location, De Smet), Laura Ingalls Wilder lived in De Smet. So, the answer is \{De Smet\}.
\\

\bottomrule
    \end{tabular}
    \caption{Detailed KGQA prompts for modules of \readi. }
    \label{tab:prompt_list}
\end{table*}

\begin{table*} 
    \setlength{\abovecaptionskip}{0.1cm}
    \setlength{\belowcaptionskip}{-0.5cm}
    \centering

    \begin{tabular}{p{15.5 cm}}
    \toprule
    \textbf{Prompts for reasoning path generation} \\You should predict the needed header and rows in a table for the question.\\
        \textbf{Demonstration Example}\\
Question:
what was the last year where this team was a part of the usl a-league? \\
| year | division | league | regular season | playoffs | open cup | avg. attendance |\\
| -- | -- | -- | -- | -- | -- | -- |\\
| 2001 | 2 | USL A-League | 4th, Western | Quarterfinals | Did not qualify | 7,169 |\\
Thought:\\
First, according to headers and example rows, I need the years the team is in usl a-league league and return the latest year, so I need headers "year" and "league".\\
Second, I need to constrain "league" = "usl a-league" to know the years of this team as part of the "usl a-league", so I need {"league": ["usl a-league"]}.\\
Chosen Headers: ["year", "league"]\\
Constrains: {"league": ["usl a-league"]}\\
\midrule
    \textbf{Prompts for reasoning path editing} \\
    There are some mistakes in your previous header or constrains of a question.
Follow the given feedback, fix your mistakes and give the correct header and constrains.\\
\textbf{Demonstration Example}\\
Question: what was the last year where this team was a part of the usl a-league?\\
| year | division | league | regular season | playoffs | open cup | avg. attendance |\\
| -- | -- | -- | -- | -- | -- | -- |\\
| 2001 | 2 | USL A-League | 4th, Western | Quarterfinals | Did not qualify | 7,169 |\\
Wrong Answer: \\
Chosen Headers: ["year", "team"]\\
Constrains: {"Team": ["usl a-league"]}\\
Feedback:\\
1. Header ['team'] not in candidate Headers. You can only choose headers from ["year", ..."avg. attendance"].\\

Thought:
First, previously I chose headers "year" and "yeam", but "team" is not in Header list. Following the feedback, I need the team in "league"="usl a-league" , so I need headers "year" and "league".\\
Second, I need to constrain "league" = "usl a-league".\\
Chosen Headers: ["year", "league"]\\
Constrains: {"league": ["usl a-league"]}\\
\midrule
    \textbf{Prompts for QA reasoning} \\

    You should output the answer of question based on a table.\\
Output your answer in the last line as "Answer: ['your answer']"!\\
    \textbf{Demonstration Example}\\
Question: what was the last year where this team was a part of the usl a-league?\\
Table:\\
Headers: league, year\\
item 1: (league, usl a-league); (year, 2001)\\
item 2: (league, usl a-league); (year, 2002)\\
item 3: (league, usl a-league); (year, 2003)\\
item 4: (league, usl a-league); (year, 2004)\\
Thought:\\
First,  I know the years the teams is a part of usl a-league are 2001, 2002, 2003 and 2004 from the items in Table.\\
Second, I calculate the last year is 2004, so the answer is ['2004'].\\
Answer: ['2004']\\

\bottomrule
    \end{tabular}
    \caption{Detailed TableQA prompts for modules of \readi. }
    \label{tab:prompt_list_table}
\end{table*}

\end{document}